\pdfoutput=1

\documentclass[11pt]{article}

\usepackage[]{acl}

\usepackage{times}
\usepackage{latexsym}

\usepackage{microtype}
\usepackage{graphicx}
\usepackage{paralist}
\usepackage{multirow}
\usepackage[export]{adjustbox}
\usepackage{booktabs}
\usepackage{nicematrix}
\usepackage{xcolor}
\usepackage{colortbl}
\usepackage[outline]{contour}
\usepackage{enumitem}
\usepackage{siunitx}
\usepackage{subcaption}
\usepackage{comment}
\usepackage{amsmath}

\usepackage{latexsym}
\usepackage{amsfonts}
\usepackage[utf8]{inputenc}
\usepackage[T1]{fontenc}
\usepackage{bbm}
\usepackage{amssymb}

\newcommand{\ku}{$^1$}
\newcommand{\is}{$^2$}
\newcommand{\com}{$^3$}

\title{Understanding Retrieval Robustness for \\Retrieval-Augmented Image Captioning}

\author{Wenyan Li\ku, Jiaang Li\ku, Rita Ramos\is, Raphael Tang\com, Desmond Elliott\ku \\
{\ku}Department of Computer Science, University of Copenhagen \\
{\is}INESC-ID, Instituto Superior Tecnico, University of Lisbon {\com}Comcast Applied AI\\
\texttt{\{weli, jili, de\}@di.ku.dk} \\
  \texttt{ritaparadaramos@tecnico.ulisboa.pt} \\
  \texttt{raphael\_tang@comcast.com}}

\begin{document}
\maketitle

\begin{abstract}
Recent advances in retrieval-augmented models for image captioning highlight the benefit of retrieving related captions for efficient, lightweight models with strong domain-transfer capabilities. While these models demonstrate the success of retrieval augmentation, retrieval models are still far from perfect in practice:  the retrieved information can sometimes mislead the model, resulting in incorrect generation and worse performance. In this paper, we analyze the robustness of a retrieval-augmented captioning model \textsc{SmallCap}. Our analysis shows that the model is sensitive to tokens that appear in the majority of the retrieved captions, and the input attribution shows that those tokens are likely copied into the generated output. Given these findings, we propose to train the model by sampling retrieved captions from more diverse sets. This decreases the chance that  the model learns to copy majority tokens, and improves both in-domain and cross-domain performance.

\end{abstract}

\section{Introduction}
\label{sec:intro}
\begin{figure}[t]
        \includegraphics[width=\columnwidth]{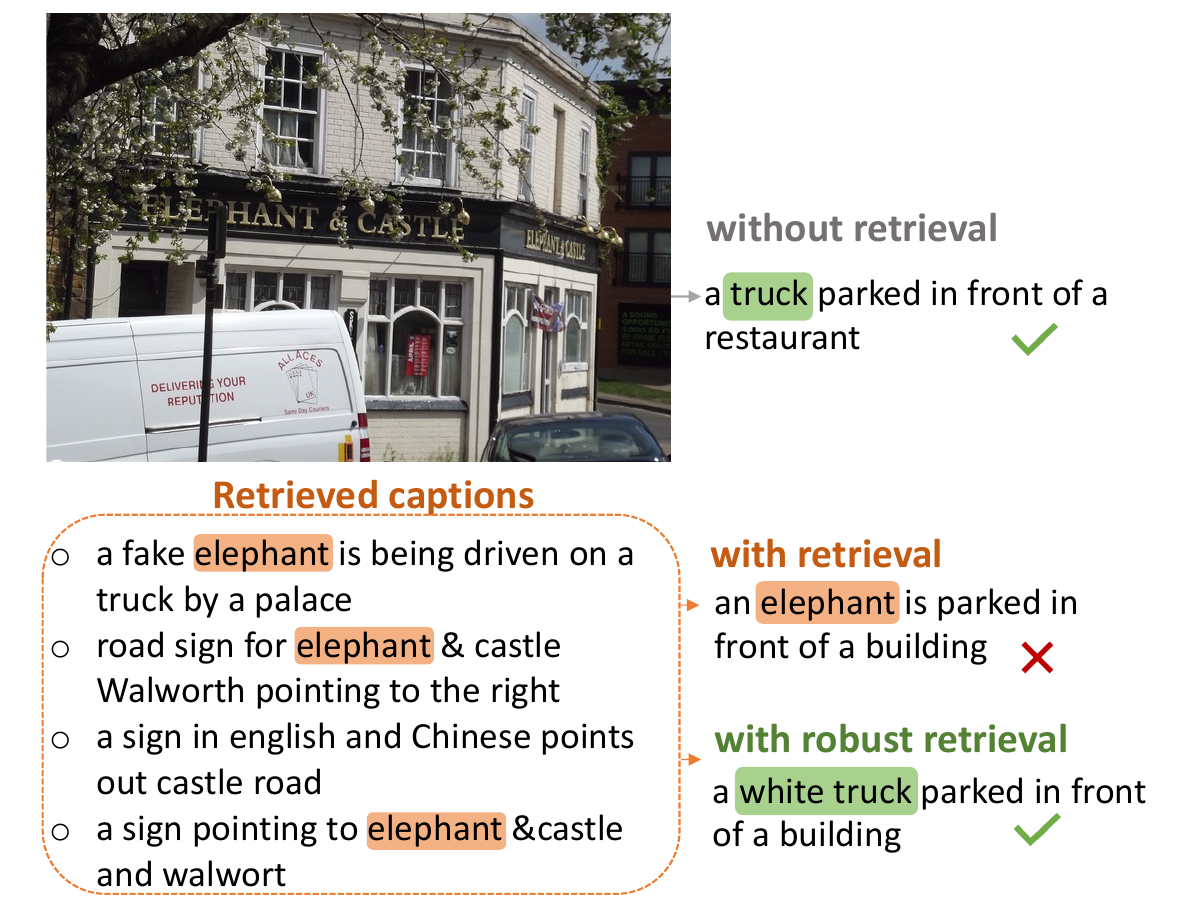}
        \caption{Comparison of generated image captions that are predicted without retrieval, misled by retrieval, and predicted with a more retrieval-robust model. The retrieval-augmented model generates the token ``elephant'', which appears in 3/4 of the retrieved captions.\label{fig:ex}}
    \end{figure}

Recent retrieval-augmented image captioning models have shown success in strong image captioning performance while reducing model parameters by retrieving related captions for a given image~\citep{ramos2023smallcap,rag-transformer,yang2023re}. These models use retrieved information as additional context besides the input image.
However, similar to retrieval-augmented language models \citep{ret-robust23}, image captioning models enhanced with retrieval can sometimes be misled by irrelevant information.
For example, in Figure~\ref{fig:ex} the captioning model is misled by the token ``elephant'' in the retrieved captions, and generates captions that do not match the given image.

For retrieval-augmented language models, \citet{ret-robust23} have studied the cases where retrieval misled the model prediction, and address this problem with a retrieval-robust LLM by continuous training with synthetic data for question answering tasks. However, in their approach, the retrieval system returns only one passage at each step. Considering that LLMs can be sensitive to the order of prompts~\cite{lu-etal-2022-fantastically}, the robustness of using multiple retrieved results has not been fully studied. Evaluating and improving the robustness of retrieval-augmented image captioning models remains under-explored, specifically when the model is augmented with multiple retrieved results.

To bridge this gap in the literature, we study the robustness of the \textsc{SmallCap} retrieval-augmented captioning model~\citep{ramos2023smallcap}. By the definition of retrieval robustness proposed in~\citet{ret-robust23}, retrieved context should boost model performance when relevant, and should not adversely affect it when irrelevant. We thoroughly examine the robustness of
the model with regards to the order of the retrieved captions, and the relevance of the retrieved content. We also present a novel analysis of model behaviour based on \textit{majority voting}, supported by input attribution and attention analyses to investigate how the retrieved tokens influence the model generation. And finally, inspired by~\citet{hoang2022improving}, we propose to sample the retrieved captions from a larger list during training to prevent the model from overfitting to the top relevant captions. Our evaluation shows improved model robustness and better out-of-domain generalization. 

The main findings of this paper are: \textbf{1)} We study the robustness of an existing retrieval-augmented captioning model \textsc{SmallCap} and find it is not robust to processing randomly retrieved content. \textbf{2)} We identify that tokens that frequently occur in the retrieved captions, i.e. majority tokens, have high attribution scores with regard to the tokens generated by the model. This phenomenon suggests heightened sensitivity and copying. \textbf{3)} Training with sampled retrieved captions from a larger list instead of with fixed top-k relevant captions improves model robustness, yielding better generalization and out-of-domain performance.\footnote{We release the code at \url{https://github.com/lyan62/RobustCap}}

\section{Related Work}
\label{sec:related}

\paragraph*{Robustness of retrieval-augmented models.}
Retrieval-augmented generation (RAG) involves enhancing the generation process by incorporating retrieved information from an external datastore as additional context to the input \cite{rag}. RAG models have shown to improve performance across a variety of NLP tasks \cite{mialon2023augmented}. 
However, RAG models can overly rely on retrieved information, resulting in inaccurate generation when the retrieved context is flawed \cite{yan2024corrective,ret-robust23}.

Recent efforts aim to enhance RAG model robustness against misguided or hallucinated generations. One approach involves filtering retrieved content \citep{wang2023learning, ret-robust23, racm3, yan2024corrective, asai2023self} by applying or training an additional evaluator. Another direction focuses on improving robustness during the training of the generation model itself. Specifically, for retrieval-augmented question answering with large language models, \citet{ret-robust23} propose continued training with a synthetic dataset that contains both relevant and irrelevant context, while \citet{cuconasu2024noise} suggests incorporating irrelevant documents. In retrieval-augmented translation, robustness is improved through shuffling retrieved translations \citep{hoang2022improving}, ensemble model decoding \citep{hao-etal-2023-rethinking}, and controlled interactions between source and retrieved translations \citep{hoang-etal-2023-improving}.

\paragraph{Retrieval-augmented image captioning.}
\label{para:raic}
Image captioning is the task that describes the visual contents of an image in natural language \cite{xu2015show, osman2023survey}. Recent studies have integrated RAG into this field. \citet{rag-transformer} and \citet{zhou-long-2023-style} experimented with retrieving similar or style-aware images before generating captions. \citet{li2023evcap} introduced a lightweight image captioning model that utilizes retrieved concepts. More related to our work, \citet{ramos-etal-2023-retrieval} developed end-to-end encoder-decoder models that attend to both the image and retrieved caption embeddings. 

In particular, the \textsc{SmallCap} model \citep{ramos2023smallcap}, presenting retrieval augmentation in image captioning could reduce trainable parameters and adapt to out-of-domain settings. The model utilizes frozen unimodal models, incorporating a pre-trained encoder and decoder connected by trainable cross-attention layer.

However, it still remains unclear how retrieved captions influence the generation of captions in retrieval-augmented image captioning, especially concerning visual inputs. Additionally, the evaluation and enhancement of the robustness of these models are still under-explored.

\section{Robustness of Retrieval-Augmented Image Captioning}
\label{sec:robust-eval}
To evaluate the robustness of the \textsc{SmallCap} retrieval-augmented caption model \cite{ramos2023smallcap}, we conduct controlled experiments 
 and observe its resilience to changes in (1) the order of the retrieved captions and (2) the content relevance of the retrieved captions.

    \begin{figure}[t]
    \centering
        \includegraphics[width=0.48\textwidth,trim={3mm 5mm 2mm 2mm},clip]{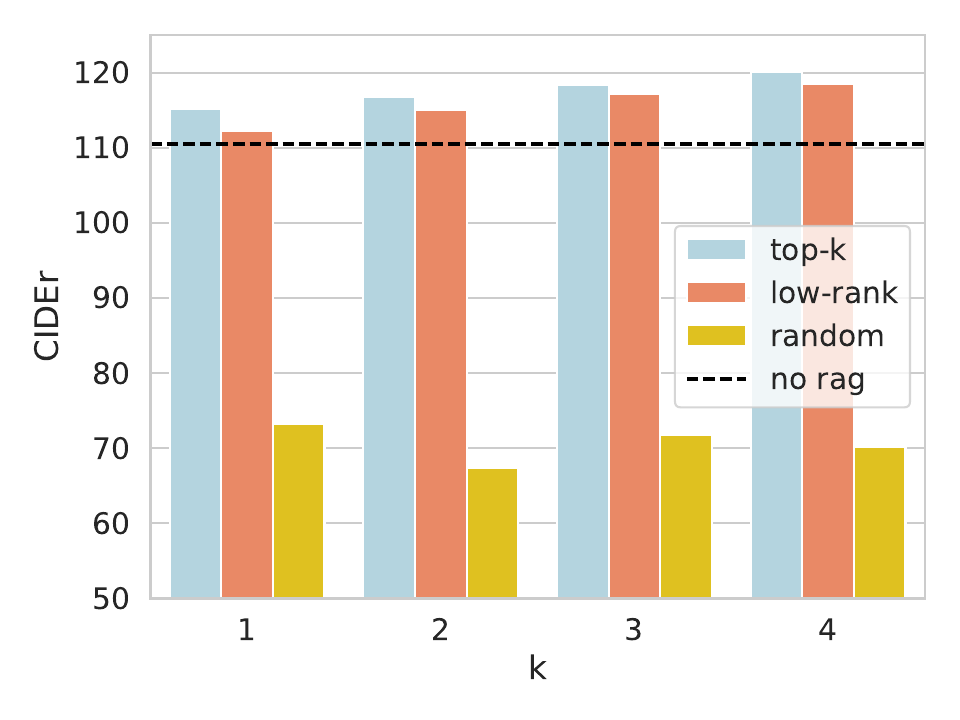}
        \caption{CIDEr evaluation of \textsc{SmallCap} on the COCO validation set using the top-$k$, low(er)-ranked, randomly retrieved captions, against a baseline without retrieval augmentation. Performance drops by up to 50\% when using randomly retrieved captions compared to baseline, suggesting that the model is not robust.\label{fig:content-sensitivity}}
    \end{figure}

\subsection{Robustness Evaluation}
For a given image, \textsc{SmallCap} is augmented with a sequence of $k$ retrieved captions that are combined into an input for the language model decoder: \textit{``Similar images show $cap_{1}, cap_{2}, ..., cap_{k}$. This image shows ...''}. The retrieved captions are obtained through image-to-text retrieval using CLIP embeddings~\citep{clip}, and are sorted according to their relevance, i.e., cosine similarity. From the sorted retrieved captions, we retain the most similar captions as the retrieval list. In this regard, the top-$k$
 retrieved candidates are the first $k$ captions in the list, and the low-ranked captions are the last-$k$ captions in the list. \textsc{SmallCap} uses the top-$k$ retrieved captions in the prompt by default.

\paragraph*{Context order.}
\label{subsec:order}
    When prompting the model to generate a caption for a given image, we can change the order of the retrieved captions by \textbf{permuting} or \textbf{reversing} them. We evaluate the effect of the order changes in two settings: one with a model trained using the top-$k$ retrieved captions (default), and another that is also trained with permuted or reversed retrieved captions.\footnote{For the model trained with default order---top four captions, we use the pretrained checkpoints from HuggingFace: \url{https://huggingface.co/Yova/SmallCap7M}, \url{https://huggingface.co/Yova/SmallCapOPT7M}} 

\paragraph*{Content relevance.}
\label{subsec:content} 
    To evaluate how robust the model is towards noise in the retrieved captions, we are curious to see how the model performs when (1) captions are randomly retrieved, i.e. likely to be irrelevant for the given image (2) only low-ranked retrieved captions are available. Here the randomly retrieved captions are those retrieved with another image. For low-ranked captions, we take the lowest-ranked $k$ captions from the retrieval list that consists of top seven relevant captions. 

\subsection{Experimental Setup} In the experiments, we set $k=4$ as it has been demonstrated as the optimal number of captions by~\citet{ramos2023smallcap}. We evaluate \textsc{SmallCap} models with both OPT-350M \cite{zhang2022opt} and GPT-2 \cite{gpt2} as the decoder models. For the image encoder, we use ResNet-50x64~\cite{he2016deep} and CLIP-ViT-B/32~\cite{clip} as the retrieval encoder. We keep the same model setting in the following sections unless stated otherwise.

\paragraph*{Data and metrics}
\label{para:order-data}
We first evaluate the robustness of \textsc{Smallcap} on COCO validation set for \textit{in}-domain evaluation. Then we evaluate on NoCaps~\citep{agrawal2019nocaps}, which contains \textit{In, Near} and \textit{Out}-of-domain data, and serves as a challenging dataset designed to assess the generalization capabilities of models trained on COCO. For both datasets we use the validation set experimenting with different number of retrieved captions, i.e. different $k$ values. We report peformance using CIDEr score~\citep{cider}. 
    
\begin{table}[t]
\centering
\renewcommand{\arraystretch}{1.1}
     \begin{tabular}{llrrrr}
        \multicolumn{2}{c}{Retrieval Order} & \multicolumn{3}{c}{LM Backbone}\\
        \midrule
        Train & Eval & \multicolumn{1}{c}{GPT-2} & & \multicolumn{1}{c}{OPT} \\
        \midrule
        \multirow{3}{*}{default}
        & default & 116.4 & & 120.3 \\
        & permute & 116.2 & & 120.1 \\
        & reverse & 115.8	& & 119.7 \\
        \midrule
        permute & permute & \textbf{117.2} & & \textbf{120.4} \\
        reverse & reverse & 116.4 & & \textbf{120.7}	\\ 
        \bottomrule
        \end{tabular}
    \caption{CIDEr evaluation on the COCO validation set with GPT-2 and OPT variants of \textsc{SmallCap} when manipulating the order of the top-k retrieved captions. } \label{tab:order-coco}
\end{table}

\subsection{Order Robust but Content Sensitive}

    \paragraph{Order robust.} From the results in Table~\ref{tab:order-coco} and Table~\ref{tab:order-nocaps}, we observe that \textsc{SmallCap} is indeed robust to the order of the retrieved texts. Permuting the order of the captions during training and evaluation show $1$ CIDEr point improvement for COCO \citep{lin2014microsoft} and $2-3$ CIDEr score increase for NoCaps \citep{agrawal2019nocaps}. This indicates that if multiple captions are used for augmentation, then permuting their order helps. 

    \paragraph{Content sensitive.}
    \label{para:content-sensitive}
    Figure~\ref{fig:content-sensitivity} shows that when using randomly retrieved captions instead of the top-$k$ most relevant captions, performance drops drastically compared to the no-retrieval baseline.\footnote{Here the top and low ranked captions are obtained from a list of top-seven captions retrieved captions ordered by their cosine similarity to the image embedding.} This implies that \textsc{smallcap} lacks resilience to noise in the retrieved captions, and the irrelevant context has the potential to mislead the model, resulting in inaccurate predictions. When prompting with low-ranked retrieved captions, while performance slightly decreases, the retrieval-augmented model still outperforms the one without retrieval.

\begin{table}[t]
    \resizebox{\linewidth}{!}{
     \begin{tabular}{lllrrrrrr}
        \multicolumn{2}{c}{Retrieval Order} & & \multicolumn{5}{c}{LM Backbone}\\
        \midrule  
        & & \multicolumn{3}{c}{GPT-2} & \multicolumn{3}{c}{OPT} \\
         \cmidrule(rl){3-5} \cmidrule(rl) {6-8}
        Train & Eval & In & Near & Out & In & Near & Out \\
        \midrule
         \multirow{3}{*}{default} & default& 80.1 & 79.4 &	69.6 & 91.0 & 84.4 & 76.3 \\
           & permute & \textbf{81.6} & \textbf{79.8} & 68.5 & \textbf{92.5} & \textbf{84.5} & 75.8 \\
           & reverse & 80.2 & 79.3 & 68.4 & \textbf{92.0} & 84.4 & \textbf{76.6}\\
        \midrule
        permute & permute &  \textbf{81.5} & 79.7 & \textbf{69.8} & \textbf{94.2} & 84.0 & \textbf{79.4} \\
        reverse & reverse & \textbf{80.4} & \textbf{80.1} & 68.4
        & \textbf{92.5} & \textbf{85.6} & 75.9\\
        \bottomrule
        \end{tabular}
        }
    \caption{Evaluation on NoCaps using CIDEr score with the GPT-2 and OPT variants of \textsc{SmallCap} when manipulating the order of the top-$k$ retrieved captions.} \label{tab:order-nocaps}
\end{table}

\section{Majority Tokens Explain Behavior}
\label{sec:mv} %

To better understand how each token of the retrieved content relates to the observed sensitivity discussed in the previous section, we hypothesize that the model is driven by the presence of majority tokens. In other words, when the model is prompted with retrieved captions, we assume that the predicted tokens are influenced by the tokens that appear in the majority of the retrieved captions. To test this assumption, we propose a majority voting analysis, followed by input attribution, and an attention analysis of the model behavior.

\subsection{Majority Tokens}
\label{subsection:mv_calc}
We first introduce the definition of majority tokens.
Let $R=[T_1, \cdots, T_n]$ represent a retrieved caption $R$, which contains a sequence of $n$ tokens. For a given image, we assume that a total of $K$ retrieved captions are used in the model prompt: $R_1$, $R_2$, \dots, $R_K$. For each token $T_i$ in the set of unique tokens from the retrieved captions, we define $T_i$ as a \textit{majority token} (denoted as $T_M$) if $T_i$ appears in more than half of the retrieved captions\footnote{Note that we remove the stop words in the retrieved captions when determining the majority tokens. The stop words are filtered from the top-100 most frequent tokens in the COCO dataset, where we manually remove meaningful tokens such as ``man'', ``two'' from the list. Please see the Appendix \ref{appendix:maj_tokens} for the complete list.}, i.e., $C_{T_i} > \frac{K}{2}$ where $C_{T_i}$ is the number of retrieved captions that contains token ${T_i}$ as in Equation~\ref{eq1}:

\begin{equation}
C_{T_i} = \sum_{l=1}^{K}\mathbbm{1}[T_i \in R_k]
\label{eq1}
\end{equation}
 For a generated caption $Y=[y_1, \cdots, y_n]$ in the evaluation data, we can calculate the majority-vote probability $P_{T_M \in Y}$ as the probability of the majority token $T_{M}$ appearing in the generated caption.

We expect that the higher the value of $P_{T_M \in Y}$, the more likely it is that the model is generating captions based on the majority tokens. 

\subsection{Experimental Setup} 
\label{subsec:mv_exp}
We test our majority vote assumption with a controlled experiment. Specifically, we analyze the predictions of the model in two settings, each provided with $K=3$ retrieved captions to ensure the presence of a majority token:

\begin{description}[style=unboxed, leftmargin=0cm]
    \item[2 Good 1 Bad (2G1B):] The retrieval set contains two relevant captions and one irrelevant caption;
    \item[2 Bad 1 Good (2B1G):] The retrieval set contains two irrelevant captions and one relevant caption.
\end{description}

The assumption is that, if there is a majority voting behavior with respect to the retrieved captions, the model will copy such majority tokens to the final output. The distinction will be clear in this setting --- in the setup 2B1G, if the model is robust to the retrieved context, the model will focus more on the good caption instead of the majority tokens in the two bad captions.

We use the COCO evaluation set and the pretrained checkpoint with the OPT decoder of \citet{ramos2023smallcap} for this analysis. Good captions are obtained using the top-two and top-one retrieved captions, respectively, for a given image. Bad captions are obtained by retrieving one or two captions, respectively, from a randomly selected image.

\begin{figure*}[t]
    \centering
    \includegraphics[width=\linewidth,trim={3mm 1mm 2cm 2mm},clip]{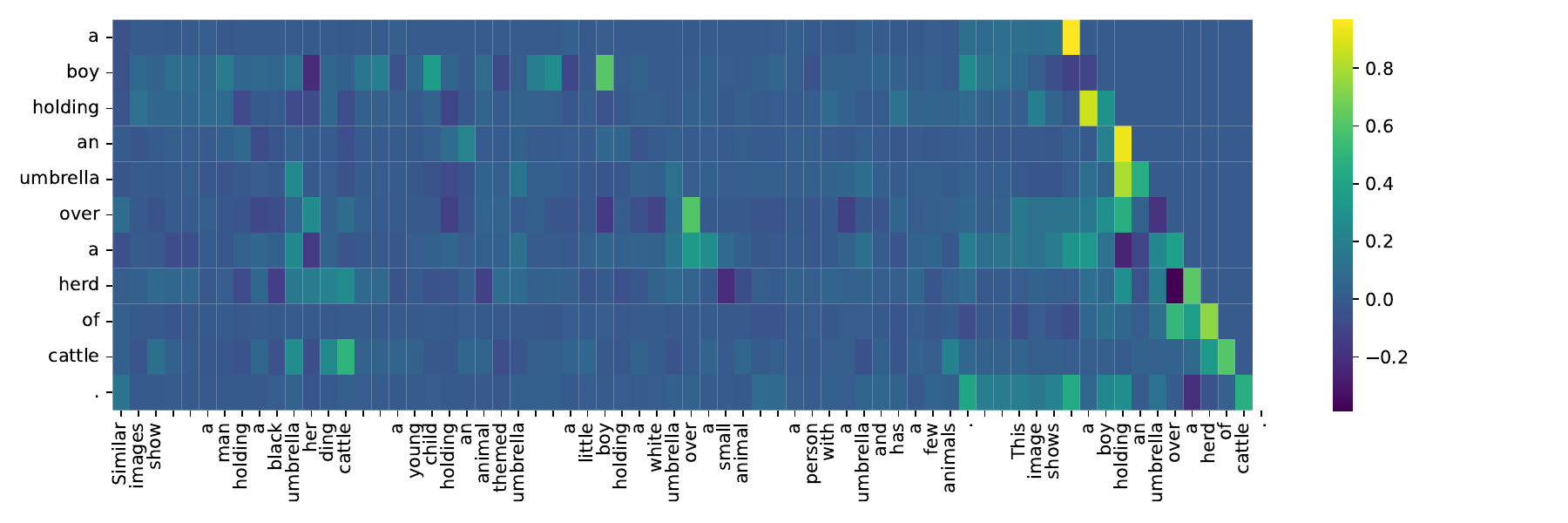}
    \caption{Input attribution for each generated token (y-axis). The brighter the color, the more greater the attribution from the input token. We observe high attribution scores to ``umbrella'', ``boy'', ``cattle'', and ``over''. \label{fig:attr-ex}}
\end{figure*}

\paragraph{Results.} %
We find that the probability of majority vote in the 2G1B setting is 86.47\%. This high probability suggests that the majority tokens in the good captions could be being used to guide the model generation. In the 2B1G setting, the model is much less likely to generate majority tokens from the bad captions, indicating some robustness in not always following them. However, 20.84\% of the time, the model can still be misled by their appearance, resulting in the majority tokens being copied into the model output.

\subsection{Input Attribution with Integrated Gradients}
\label{subsec:attribution}

To better understand the role of majority tokens in model generation, we use integrated gradients~\citep{ig} for input attribution analysis. This enables us to examine the influence of each individual token in the retrieved captions on the model prediction.

\paragraph*{Attribution visualization.} 
Figure~\ref{fig:attr-ex} shows an example of an attribution visualization, where the attribution score of each input token (x-axis) is computed at each generation step (y-axis). Bright color cells correspond to high attribution to the input token. High attribution scores to the same tokens seen in the retrieved captions may indicate copying. Negative attribution scores are observed at contradicting tokens observed in the retrieved captions to the current generation. Negative scores are observed at token ``her'' when model is predicting the token ``boy'' and at token ``small'' when predicting ``herd''. Additional input attribution visualizations can be found in Appendix~\ref{appendix:attribution_vis}.

\begin{figure}[t]
    \includegraphics[width=0.5\textwidth,trim={3mm 2mm 2mm 2mm},clip]{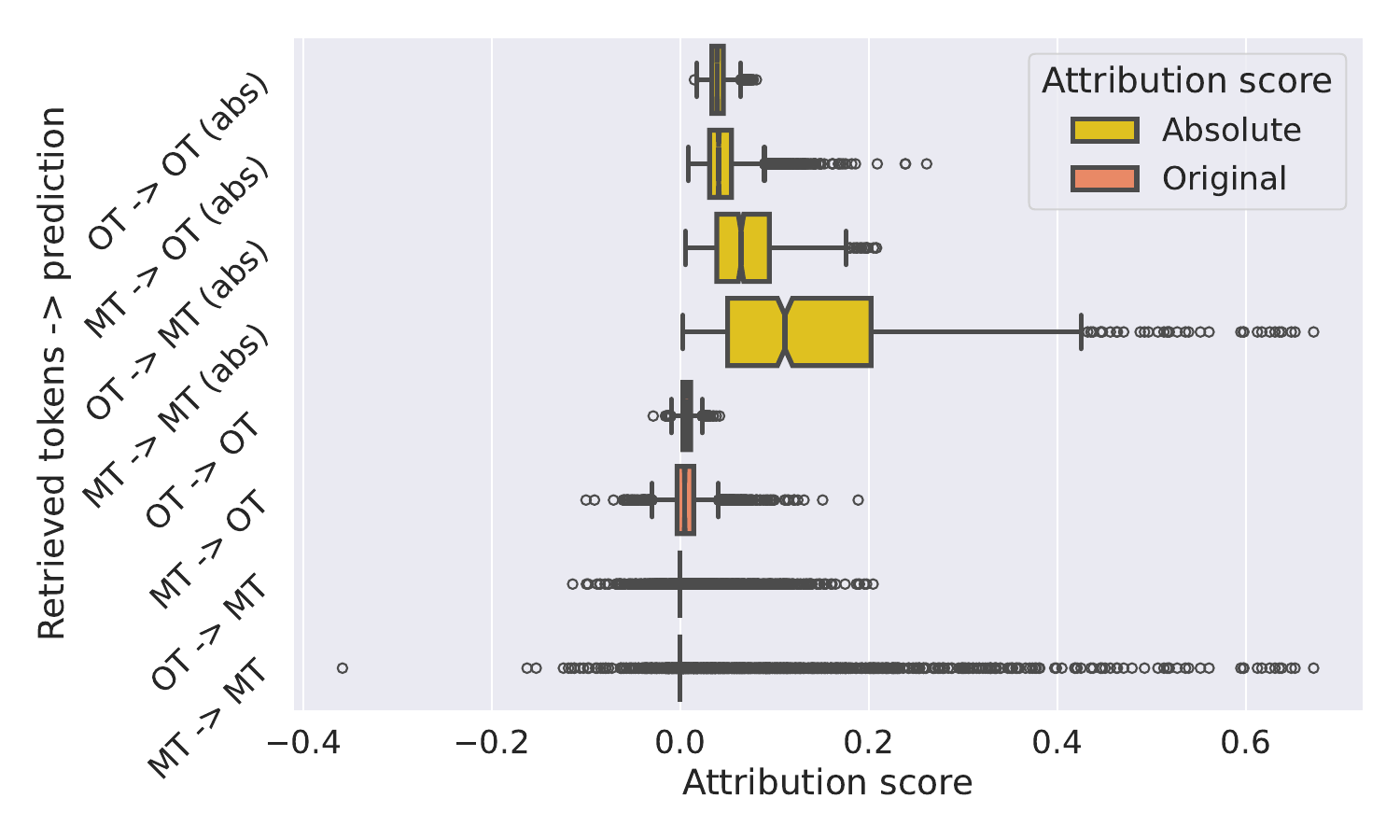}
    \caption{Pairwise average attribution score between retrieved and generated tokens in the 2B1G setup. MT: majority tokens in the retrieved captions. OT: all other tokens. The larger pairwise attribution values shows that the majority tokens have a larger impact during generation than the other tokens in the retrieved captions. \label{fig:pairwise-attr}}
\end{figure}

\paragraph*{Quantitative analysis.}
We also quantitatively analyze the impact of majority tokens by calculating pairwise attribution scores between tokens in retrieved captions and those predicted by the model. Higher attribution values suggest greater sensitivity to the input token~\cite{ancona2018towards}. Figure~\ref{fig:pairwise-attr} shows the distribution of the pairwise attribution scores for the 2B1G setup. It is clear that the model is sensitive to the majority tokens, especially when the generated token exists in the retrieved captions. Such behavior indicates weak robustness: we would not expect a robust model to be distracted by the tokens from the two irrelevant retrieved sentences at inference time. To better visualize the impact, we show distribution of original attribution values and the absolute values~\cite{ancona2018towards} across all evaluation samples.

\subsection{Attention and Model Behavior}
\label{subsec:attention}
Finally, we visualize the self-attention and cross-attention to locate the heads and layers in the \textsc{SmallCap}-OPT125M model that may contribute to the majority voting behaviour when generating a caption. This is crucial because all interactions between captions (self-attention) and images (cross-attention) take place in this stage. 

\begin{figure}[t]
    \centering
    \includegraphics[width=1\linewidth]{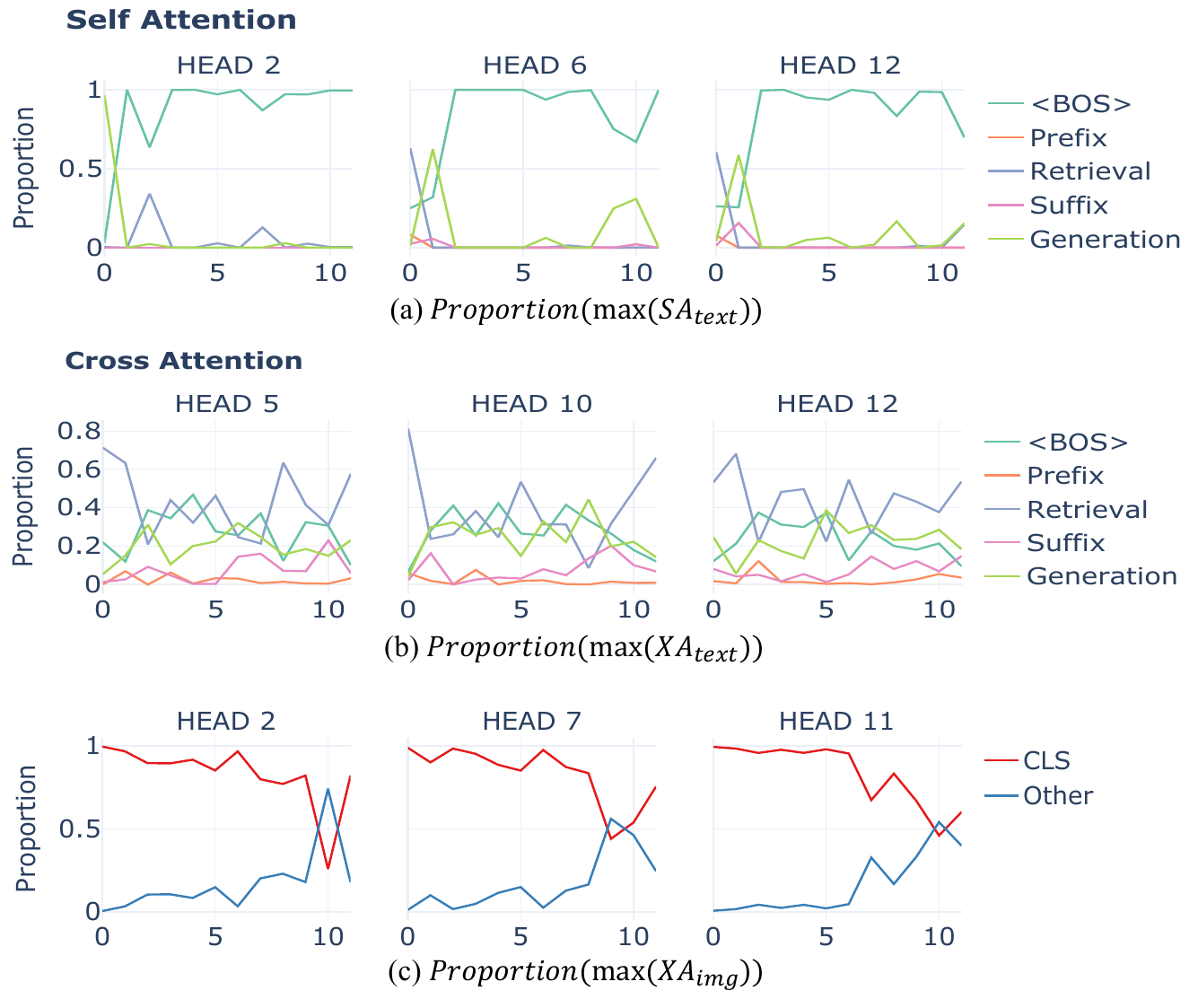}
    \caption{Statistics of all maximum attention scores' distribution across different layers and heads from self and cross attention. $XA$ denotes cross attention, while $SA$ signifies self-attention. $img$ represents the distribution of maximum attention scores across image patches, whereas $text$ pertains to the distribution of maximum attention scores across text tokens. }
    \label{fig:opt-self-cross-attentions}
\end{figure}

\paragraph{Distribution of max attention occurrence.}
We partition the text input prompt into five distinct segments: begin of the sentence token (\textit{<BOS>}), prompt tokens before retrieved $k$ captions (\textit{prefix}), i.e. ``Similar image shows'', the retrieved captions (\textit{retrieval}) $cap_1, \cdots cap_k$, prompt tokens before generation (\textit{suffix}), i.e. ``This image show'', and the \textit{generation} itself. For image patches, we segment them into two pieces -- the \textsc{CLS} output embedding, and the set of patch output embeddings.

Let $S_n$ denote the sets of indices, where $n = 1, 2, \ldots, 5$ for five segments. For the text input, each segment $S_n$ contains the indices of the tokens in each segment. To track the occurrence of max attention values in $S_n$,
we define the indicator function $\mathbbm{1}[I_n(i, j)]$ as follows:
\begin{equation}
\mathbbm{1}[I_n(i, j)] = \begin{cases}
1 & \text{if } \arg\max_z Att(j,z)_i \in S_n \\
0 & \text{otherwise}
\end{cases},
\label{eq:indicator}
\end{equation}
where $\arg\max_z Att(j,z)_i$ is the index of the input with the maximum attention score for sample $i$.

For self-attention between the textual tokens, $Att(j,z)$ represents the attention score between the $j^{th}$ generated token and the $z^{th}$ text context token, denoted as $SA_{text}(j, z)$.

For cross-attention between the decoder and the image representations, we report both a text-centric and an image-centric analysis. The text-centric analysis $XA_{text}(j, z)$ measures the attention between the $j^{th}$ image patch and the $z^{th}$ text token, to identify which segments of the text have the highest cross-attention scores in relation to the image. In the image-centric analysis $XA_{img}(j, z)$, we measure the attention between the $j^{th}$ generated token and the $z^{th}$ image patch. We now redefine the $S_n$ notation to let $S_1$ represent the CLS output embedding, and $S_2$ represent the set of image patch embeddings, respectively. This allows $XA_{img}(j, z)$ to identify if the CLS patch embedding receives the highest cross-attention scores in relation to the generated tokens, or if it is the actual image patch embeddings.

For each analysis, $SA_{text}(j, z)$, $XA_{text}(j, z)$, and $XA_{img}(j, z)$, we calculate the proportion of occurrences of the maximum score in $S_n$ by averaging through all generated tokens for a dataset.

\paragraph{Self-attention.} %
\label{para:sa}

We gather attention scores between the generated tokens and context tokens, and categorize the distribution of the maximum scores into the five text segments (BOS, prefix, retrieval, suffix, and generation).

Figure \ref{fig:opt-self-cross-attentions}(a) illustrates the 
changes in the distribution of maximum self-attention scores in each layer of the decoder language model. 
Notably, at the initial layers, a majority of attention heads exhibit heightened focus on retrieved captions or the current context for generation. However, after the second layer, we observe an increased emphasis on the beginning of sentence token (\texttt{</s>}). This behavior is consistent with prior research on the attention mechanism of GPT-2 \cite{gpt2-analyzing}. Figures~\ref{subfig:gpt2-self-attention} and~\ref{subfig:opt-self-attention} show the behaviour for all self-attention heads in for the GPT and OPT model variants, respectively.

\paragraph{Cross-attention.}

Similar to the self-attention behaviour, we categorize the occurrence of the maximum cross-attention to the five text segments.
As shown in Figure \ref{fig:opt-self-cross-attentions}(b), in most attention heads, the cross-attention attains its maximum value between the image and the retrieved captions or between the image and the generated tokens. Figures~\ref{subfig:gpt2-cross-attention-text} and \ref{subfig:opt-cross-attention-text} show the text-centric analysis for all cross-attention heads for the GPT and OPT backbones.

Finally, we inspect whether the model focuses on the \textsc{CLS} patch or actual image patches. In Figure \ref{fig:opt-self-cross-attentions}(c), we observe that the model only pays maximum attention to the image patches in the final layers (the blue line). Figures \ref{subfig:gpt2-cross-attention-cls} and \ref{subfig:opt-cross-attention-cls} show the full results for the image-centric analysis. 

Overall, these observations show that the model attends to both modalities during the caption generation process. However the lack of strong cross-attention to actual image patches suggests that the model is misled by text prompts, even when irrelevant information is absent in the provided image.

\section{Improving Robustness to Retrieval via Sampling}
\label{sec:sample}

In order to improve the robustness of the model to potentially noisy captions, we propose to randomly sample the captions from a larger retrieval list for a given image, instead of training with only the top-$k$ retrieved captions. In this manner, the model can learn from more diverse context that includes both top- and lower-ranked captions.

\subsection{Experimental Setup}
\label{subsec:exp}
Inspired by \citet{hoang2022improving}, we experiment with two sampling methods during training to improve retrieval robustness.

\paragraph*{Sample-$k$ training.}
\label{para:training}
We sample $k$ captions randomly from the top-$N$=7 retrieved captions during training\footnote{We sample from the top-$N$=7 for alignment with the baseline; see the Appendix for an ablation on varying $N$.}. Following \citet{ramos2023smallcap}, we train \textsc{smallCap} with the OPT-350M decoder on the COCO captioning dataset~\citep{chen2015microsoft} for 10 epochs on a NVIDIA A100 40GB GPU with the default learning rate of 1e-4 and batch size of 64. We experiment with $k$ in the range of 1--4.

\paragraph*{Controlled sample-$k$ training (c-sample-$k$).}
Aiming to train the model that better distinguishes irrelevant context, we design a controlled sampling process --- selecting $k-1$ randomly from the larger list while keeping the top relevant caption of the image during training. We train the model with same hyperparameters and dataset as sample-$k$.

\subsection{Evaluation and Results}
\label{subsec:results}

In addition to the COCO and NoCaps validation set, we evaluate the \textit{Out}-domain performance of the model using VizWiz caption dataset~\citep{vizwiz} and report CIDEr scores.

\begin{table}[ht]
\centering
    \resizebox{0.9\linewidth}{!}{
     \begin{tabular}{llccc}
        \toprule   
          & & \multicolumn{3}{c}{COCO Eval} \\
          \cmidrule(lr){3-5} 
        Model & $k$ & top-$k$ & last-$k$ & random \\
        \midrule
        \rowcolor[gray]{0.95}
         top-k & 1 & 115.1 & 112.2 & 73.2 \\
         sample-k & 1 & \textbf{116.0} & \textbf{115.0} & \textbf{98.9} \\
        \arrayrulecolor{lightgray}\midrule
        \rowcolor[gray]{0.95}
        top-k & 2 & 116.8 & 115.0 & 67.4 \\
        sample-k & 2 & \textbf{117.4} & \textbf{116.8} & \textbf{84.6} \\
        \arrayrulecolor{lightgray}\midrule
        \rowcolor[gray]{0.95}
        top-k & 3 & 118.3 & 117.1 & 71.8 \\
        sample-k & 3 & \textbf{118.5} & \textbf{117.3} & \textbf{77.6} \\
        \arrayrulecolor{lightgray}\midrule
        \rowcolor[gray]{0.95}
        top-k & 4 & 120.1 & 117.1 & 70.1 \\
        sample-k & 4 & 119.2 & \textbf{118.6} & \textbf{73.1} \\
        \arrayrulecolor{lightgray}\midrule
        \rowcolor[gray]{0.95}
        c-sample-k & 4 & 119.3 & \textbf{118.9} & \textbf{72.6} \\
        \arrayrulecolor{black}\bottomrule 
        \end{tabular}
    }
    \caption{CIDEr scores when training on the top-$k$, sample-$k$ and c-sample-$k$ captions. Training by sampling the retrieved captions almost always outperforms \textsc{SmallCap} for all $k$ values. It also reduces the gap between using top-relevant and low-ranked retrieved captions. Results are averaged over three seeds. Improved scores are in \textbf{bold}.\label{tab:sample}}
\end{table}

\begin{figure*}[t]
    \centering
        \includegraphics[width=\linewidth, trim=1cm 4cm 0cm 0cm, clip]{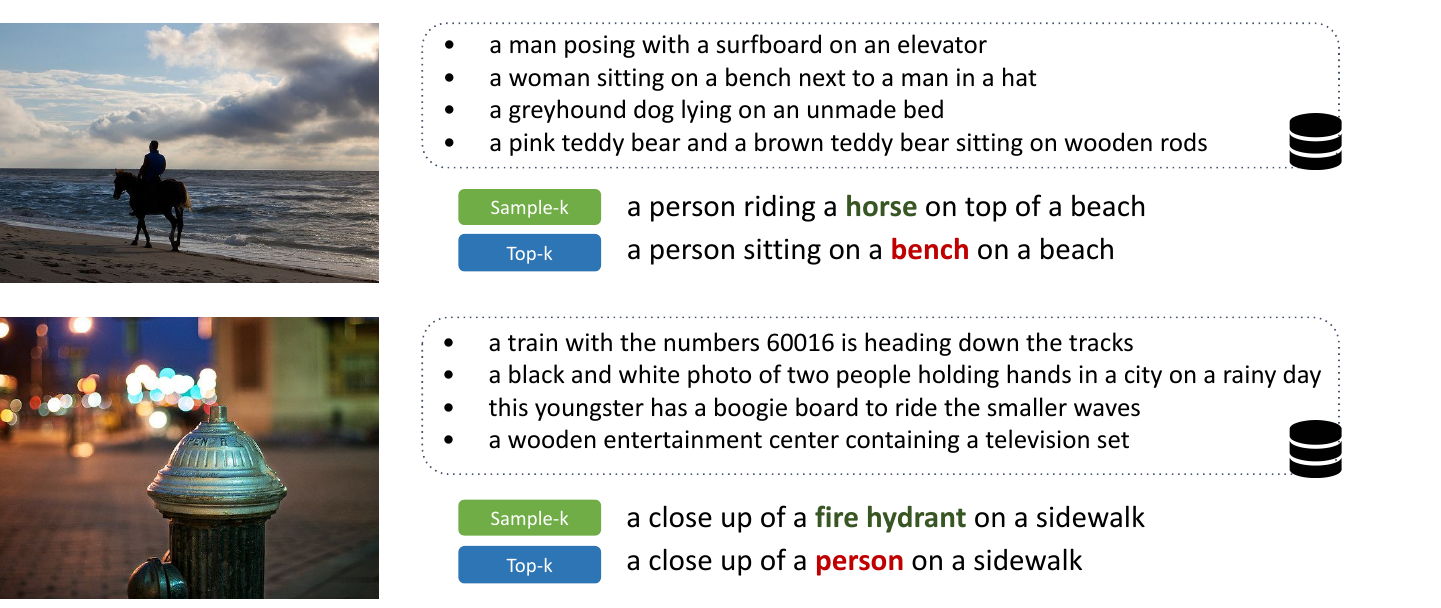}
        \caption{Qualitative examples of generated captions when \textbf{randomly} retrieving four captions for a given image using a model trained with either the Sample-k or the Top-k method.
        \label{fig:qualitative}}
\end{figure*}

\paragraph*{Sample-$k$ training improves model robustness to random retrieved captions.} 
As shown in Table~\ref{tab:sample}, incorporating sampled retrieved captions into training consistently enhances performance across various $k$ values. The improvement is particularly notable when captions are randomly retrieved, suggesting the model is now better able to ignore irrelevant context. 
If we compare across different values of $k$, sampling mitigates the model's sensitivity to the number of retrieved captions, outperforming top-$k$ training. For instance, it achieves comparable performance with a smaller $k$ value than in the case of top-$k$ training. Furthermore, the gap between using the top-$k$ vs. the last-$k$ retrieved captions is reduced with sample-$k$ training: the maximum gap is reduced from 3.0 to 1.0 CIDEr points, indicating increased model robustness, even with lower-ranked retrieved captions. Figure~\ref{fig:qualitative} and~\ref{fig:pred} show qualitative examples of the improved robustness to randomly retrieved examples.

\begin{table}[t]
\centering
    \resizebox{1\linewidth}{!}{
     \begin{tabular}{llccccccc}
        \toprule   
          & & VizWiz & \multicolumn{3}{c}{NoCaps} & \multicolumn{3}{c}{NoCaps (+Web)}
          \\ \cmidrule(rl){4-6} \cmidrule(rl) {7-9}
        Model & $k$ &  & In & Near & Out & In & Near & Out \\
        \midrule
        \rowcolor[gray]{0.95}
         top-k & 1 
         & 31.3
         & 85.0 & 74.3 &	62.3
         & 84.1 & 80.7 & 81.5 
          \\
         sample-k
        & 1 & \textbf{32.3}
        & \textbf{87.0} & \textbf{75.7}	& \textbf{63.6} &
        \textbf{87.8} &  \textbf{81.2} & 77.5 
         \\
        \arrayrulecolor{lightgray}\midrule
        \rowcolor[gray]{0.95}
        top-k & 2 & 33.7 
        & 85.0	& 74.3 & 62.3
        & 90.5 & 86.2 & 89.5 
        \\
        sample-k & 2  
         & \textbf{34.0}
         & \textbf{87.8} & \textbf{77.4}	&\textbf{67.6}	
         &\textbf{90.6}  & 85.3  & 86.7 
          \\
         \arrayrulecolor{lightgray}\midrule
         \rowcolor[gray]{0.95}
         top-k & 3 & 35.0
        & 87.4	& 79.6	& 68.3
        & 91.7 & 88.3  & 89.9 
         \\
        sample-k & 3  
         & \textbf{35.4} 
         &\textbf{88.7} & \textbf{80.3}	&\textbf{69.4}	
         &\textbf{92.6}  & 88.0  & \textbf{90.0}
         \\
         \arrayrulecolor{lightgray}\midrule
        \rowcolor[gray]{0.95}
         top-k & 4 & 35.5
        & 87.4	 & 79.6 & 68.3
        &94.2  & 89.4  & 91.2
         \\
        sample-k & 4  
          & \textbf{35.7}
         & \textbf{89.7} & \textbf{80.9}	&\textbf{71.1}	
         &\textbf{94.8}  & \textbf{89.5} &\textbf{93.1}
        \\
        \arrayrulecolor{lightgray}\midrule
         c-sample-k & 4  
        & \textbf{36.0}
         & \textbf{90.1} & \textbf{81.3}	&\textbf{71.5}	
         & \textbf{94.5} & \textbf{90.0} &\textbf{93.3}
         \\
        \arrayrulecolor{black}\bottomrule 
        \end{tabular}
        }
    \caption{Training with sampled retrieval always outperforms top-$k$ retrieval for all values of $k$ on the out-of-domain VizWiz and NoCaps datasets. The gains are smaller when using a larger datastore (+Web) but it still improves out-domain performance when retrieving more captions. Improved scores are in \textbf{bold}.\label{tab:sample-2}}
\end{table}

\paragraph*{Sampling improves cross-domain evaluation.} We also evaluate on VizWiz and NoCaps to measure cross-domain performance (Table~\ref{tab:sample-2}). This is a more realistic setting where retrieved captions are out-of-domain and could be more noisy and less relevant. The application of sampling improves across all values of $k$ for Vizwiz. On the NoCaps dataset, with the COCO datastore, sampling consistently improves near and out-domain performance, suggesting increased robustness to noisy retrieval context.  This is consistent with the benefits of sampled training demonstrated in cross-domain machine translation by \citet{hoang2022improving}. If we use a larger datastore that incorporates internet-derived captions (+Web), this consistently improves in-domain performance. Retrieval constraints are alleviated for near and out-domain samples with the larger datastore, where we see smaller gains with \text{sample-$k$}. See qualitative examples in Figure~\ref{fig:qua-nocaps} in Appendix~\ref{appendix:qua}.

\paragraph*{Controlled sampling further improves cross-domain evaluation.} Finally, on top of our best performing sample-$k$ model, controlled sample-$k$ further improves performance for both NoCaps and VizWiz. This suggests that incorporating both top-relevant and low-ranked captions during training aids the model in distinguishing irrelevant context.

\section{Discussion}
\label{subsec:discuss}
\paragraph*{Majority tokens are reliable hints during training.}

To better understand why the model relies on majority tokens during generation, we calculate the probability that majority tokens in the retrieved captions overlap with the ground truth captions ($T_{M}\in GT$), and with the predicted tokens ($T_{M}\in Pred$). Table~\ref{tab:mt-pct} shows that in 88\%--99\% of the training examples, the majority tokens in the retrieved captions are also present in the ground truth captions. This suggests that the model can develop a bias towards majority tokens due to the fact that they are so often present in the ground truth during training. This analysis also clarifies the decrease in the model's robustness as $k$ increases when randomly retrieving captions. This is because a higher $k$ only adds noise without providing useful majority tokens. The use of sampling during training exposes the model to more diverse context, which leads to a slightly increased level of selectivity.

\begin{table}[ht]
\centering
    \resizebox{0.9\linewidth}{!}{
     \begin{tabular}{lrrrr}
      & k=2 & 3 & 4 \\
        \toprule 
        \rowcolor[gray]{0.95}
        \rowcolor[gray]{0.95}
        $T_{M}\in GT_{train}$  & 88.0 & 97.5 & 99.2 \\
        \rowcolor[gray]{0.95}
        $T_{M}\in GT_{val}$  & 74.7 & 86.5 & 91.0 \\
        $T_{M}\in\text{Pred}$ & 82.8 & 93.4  & 96.7 \\
        $T_{M}\in\text{Pred (sample-k)}$& 81.9  & 93.3 & 96.6 \\
        \bottomrule
        \end{tabular}
        }
    \caption{Percentage of samples in the COCO train and validation set where the majority token of the retrieved captions are present in the ground truth compared to the percentage of their presence in prediction. \label{tab:mt-pct}}
\end{table}

    \begin{figure}[!t]
    \includegraphics[width=0.95\linewidth]{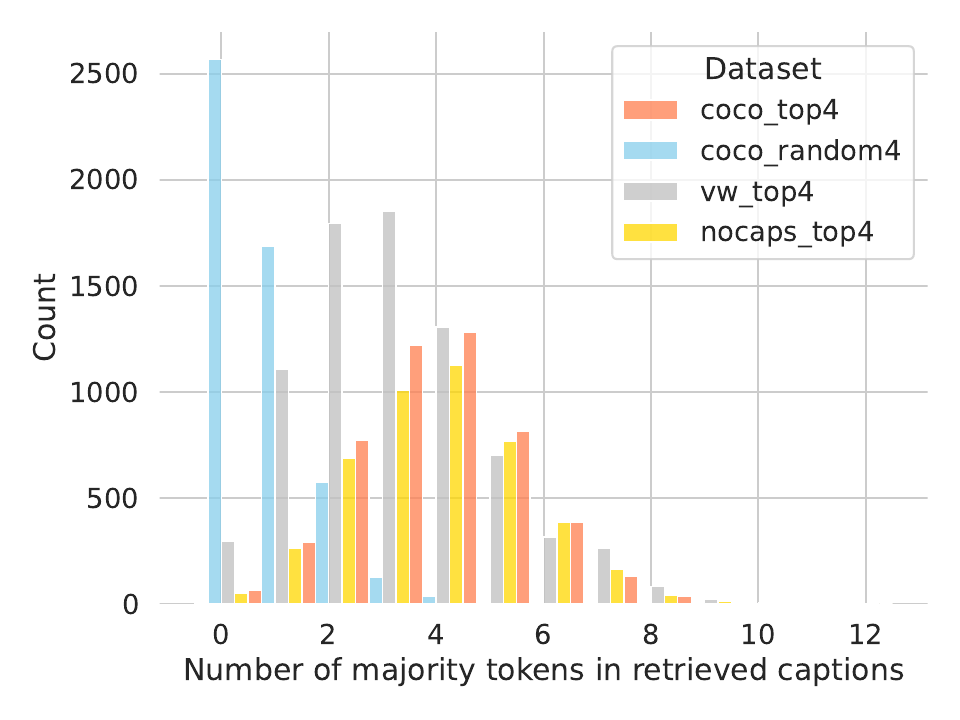}
    \caption{Distribution of number of majority tokens in the retrieved captions for the COCO, VizWis, and NoCaps evaluation datasets. For the COCO dataset, we also show the difference between retrieving the top-4 captions against four randomly selected captions.\label{fig:mt-dist}}
    \end{figure}

In Figure~\ref{fig:mt-dist}, we show the variation in the distribution of majority tokens across various evaluation datasets. When captions are randomly selected for the COCO evaluation data, there are fewer majority tokens in the retrieved captions. This presents a challenge for the model in making use of the retrieved captions, which accounts for the performance decrease shown in Figure~\ref{fig:ex}. For evaluation, with the same value of $k$, the fewer the number of majority tokens in the retrieved captions, the harder it is for the model to ``copy'' those tokens to the final output. In such scenarios, we obtain bigger improvements with the sample-$k$ training.

\section{Conclusion and Future Work}
We studied the robustness of the state-of-the-art retrieval-augmented image captioning model \textsc{SmallCap} and provide an through analysis and explanation of how retrieved captions effect the final prediction. Our exploration shows that \textsc{SmallCap} is robust to the order of the retrieved captions, but it is sensitive to retrieval noise, which has implications for using retrieval-augmented models in new domains. With extensive input attribution analysis, we show that such sensitivity is due to majority tokens in the retrieved captions. We demonstrate a more retrieval robust model can be trained with sampling methods during training. We expect that our analysis can inspire better retrieval-robust captioning models in the field. 

In the future, we will investigate whether the majority voting behaviour is exploited in other retrieval-augmented captioning models. We hope to further explore if other techniques such as token-dropping or prefix-tuning would further improve retrieval robustness.

\section*{Ethics Statement}
We acknowledge the potential risks of hallucination and biases introduced by retrieval augmentation in captioning models. Misleading tokens from the retrieved captions could cause the model to generate captions describing nonexistent entities or objects in images~\citep{liu2024survey, rohrbach-etal-2018-object}. This could have adverse effects, such as propagating systematic biases present in the datastore used for retrieval~\citep{foulds2024ragged}.

Despite the exploration in our work, we acknowledge that no system is perfect, and undesirable biases may still be present with our methods. We emphasize the need for continued research into techniques for identifying and mitigating hallucination and bias in retrieval-augmented models~\citep{foulds2024ragged, deng2024seeing}. We also stress the importance of responsible deployment, with human oversight and content moderation pipelines.

As researchers, we have an ethical obligation to be transparent about the potential risks and limitations of our work. We welcome further scrutiny and discussion around these critical issues within the research community. 

\section*{Limitations}
We evaluate the robustness of a single retrieval-augmented image captioning model in this study. Given variations in training process and model structures, the observed model behavior may be specific to our chosen model. Applying the same analysis to other models  would be useful for a deeper understanding regarding explainability and interpretation of retrieval augmented image captioning models, which we leave for future work. 

For all experiments in our study, we employ the same  CLIP-ViT-B/32 backbone as the image encoder. Investigating how model robustness varies with different visual encoders would enhance the scope of our study.

While training with sampling improves model robustness, it is intuitive that introducing more noise during training makes the task more challenging. In all our experiments, we train the model for same number of epochs as \textsc{SmallCap}, therefore it is not clear if the model would gain more robustness if trained longer. We are curious if there exists an optimal balance between training time and the level of noise exposure for achieving model robustness.

\section*{Acknowledgments}
 We thank Lei Li and the CoAStal and LAMP groups for feedback. Wenyan Li is supported by the Lundbeck Foundation (BrainDrugs grant: R279-2018-1145) and  a research grant (VIL53122) from VILLUM FONDEN. Jiaang Li is supported by Carlsberg Research Foundation (grant: CF221432). Rita Ramos is supported by the Portuguese Recovery and Resilience Plan through project C645008882-00000055 (i.e., the Center For Responsible AI), and also by Funda\c{c}\~ao para a Ci\^encia e Tecnologia (FCT), through the project with reference UIDB/50021/2020 (DOI:10.54499/UIDB/50021/2020) and the Ph.D. scholarship with reference 2020.06106.BD.

\bibliography{anthology,custom}
\bibliographystyle{acl_natbib}

\appendix
\newpage

\section{Majority Tokens\label{appendix:maj_tokens}}

\subsection{Stop words list}

In this section, we present the stop words that were filtered from the COCO dataset in the experiments described in Section \ref{subsec:mv_exp}:\\ 

[``out'', ``some'', ``of'', ``is'', ``while'', ``are'', ``with'', ``down'', ``has'', ``over'', ``the'', ``next'', ``up'', ``near'', ``several'', ``other'', ``at'', ``top'', ``from'', ``in'', ``on'', ``a'', ``there'', ``an'', ``to'', ``and'', ``her'', ``front'', ``by'', ``for'', ``his'', ``it'']

\section{More Visualization}

\begin{figure*}[t]
    \centering
    {\includegraphics[width=1\linewidth]{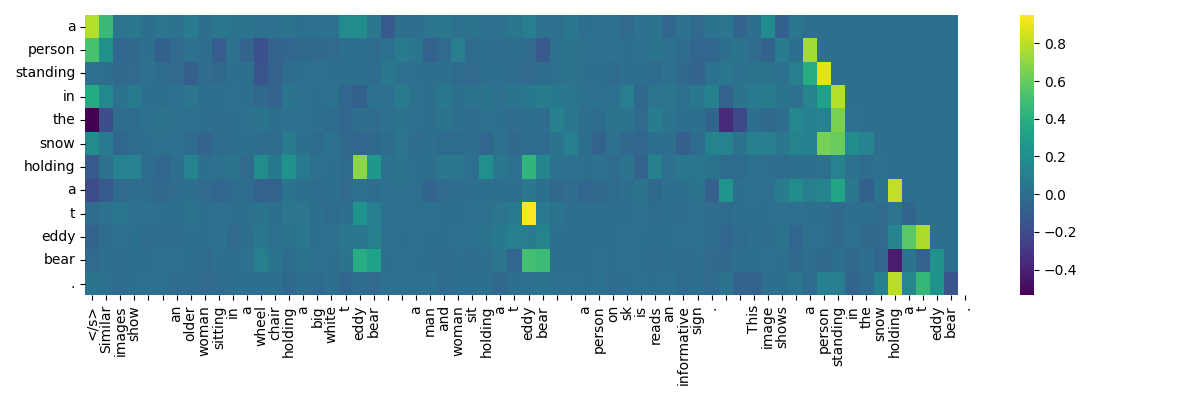}}\\
    {\includegraphics[width=1\linewidth]{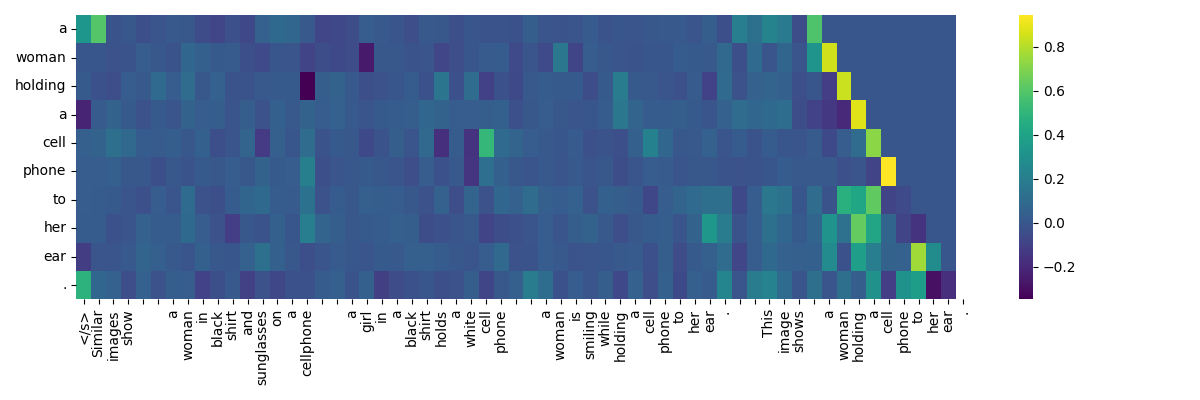}}\\
    {\includegraphics[width=1\linewidth]{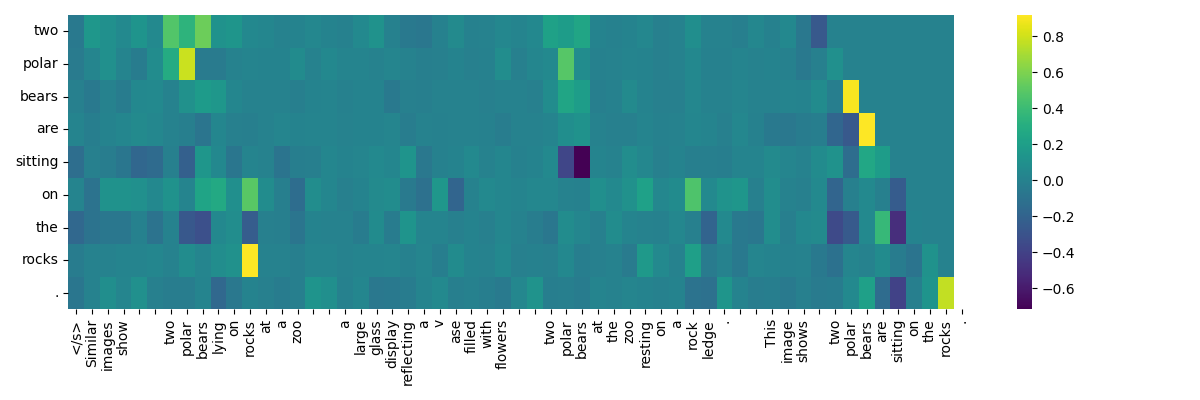}}
    \caption{Attribution visualization with few more examples. Here the model prediction is misled by the majority tokens in the 2B1G setting.}
    \label{fig:opt-attribution-vis-more}
\end{figure*}

\subsection{Input Attribution with Integrated Gradients\label{appendix:attribution_vis}}
 In Figure~\ref{fig:opt-attribution-vis-more}, we show more attribution visualization for the experiment setup 2B1G in Section~\ref{sec:mv} where high attribution scores are observed in the majority tokens and mislead the model to generate incorrect captions.

\subsection{Attention\label{appendix:attention_vis}}
\begin{figure}[ht]
    \centering
    \subfloat[\label{subfig:gpt2-self-attention}Self attention distribution.]{\includegraphics[width=1\linewidth]{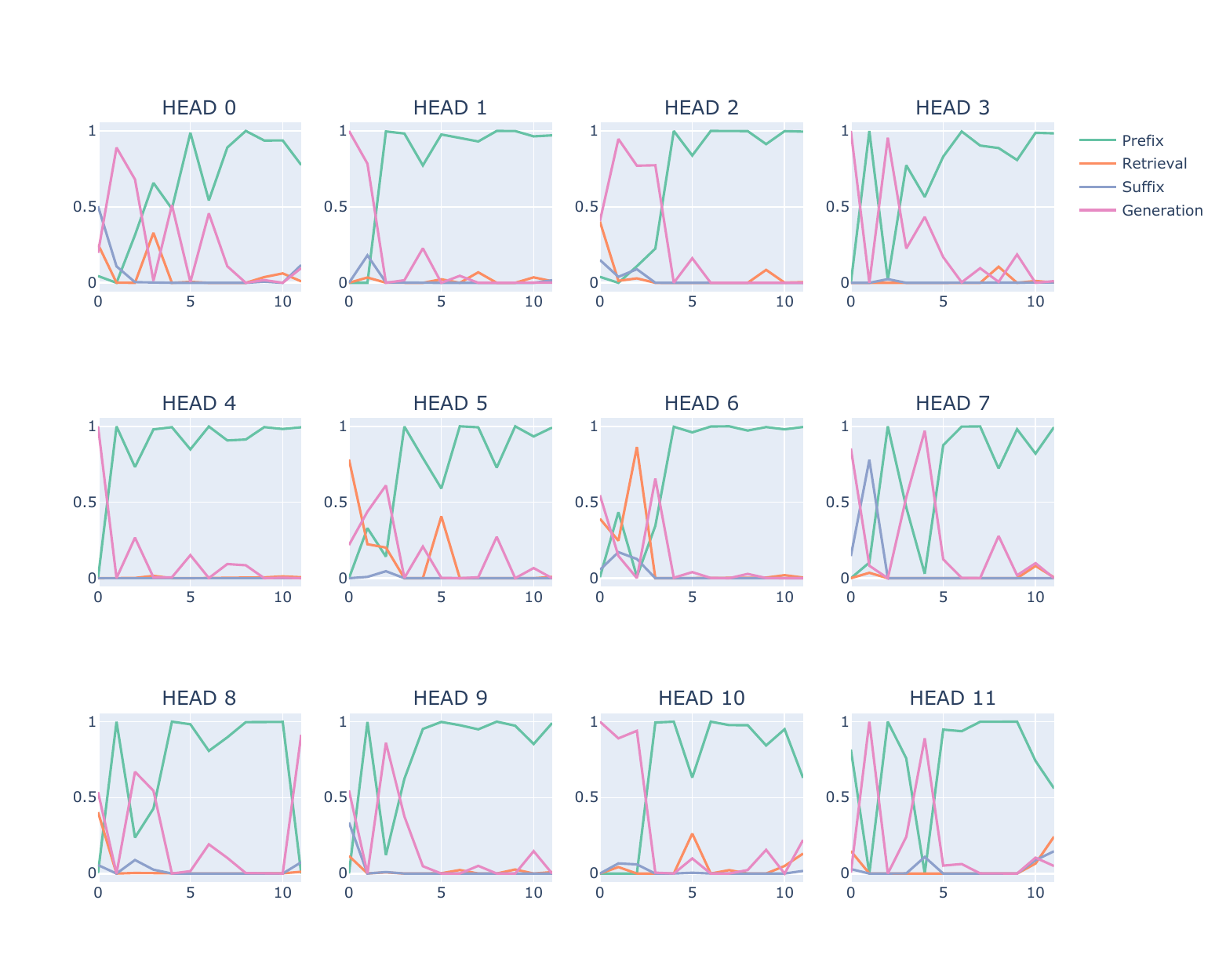}}\\
    \subfloat[\label{subfig:gpt2-cross-attention-text}Cross attention distribution. Distribution of max attention scores of the interaction between various part of text prompt and image patches.]{\includegraphics[width=1\linewidth]{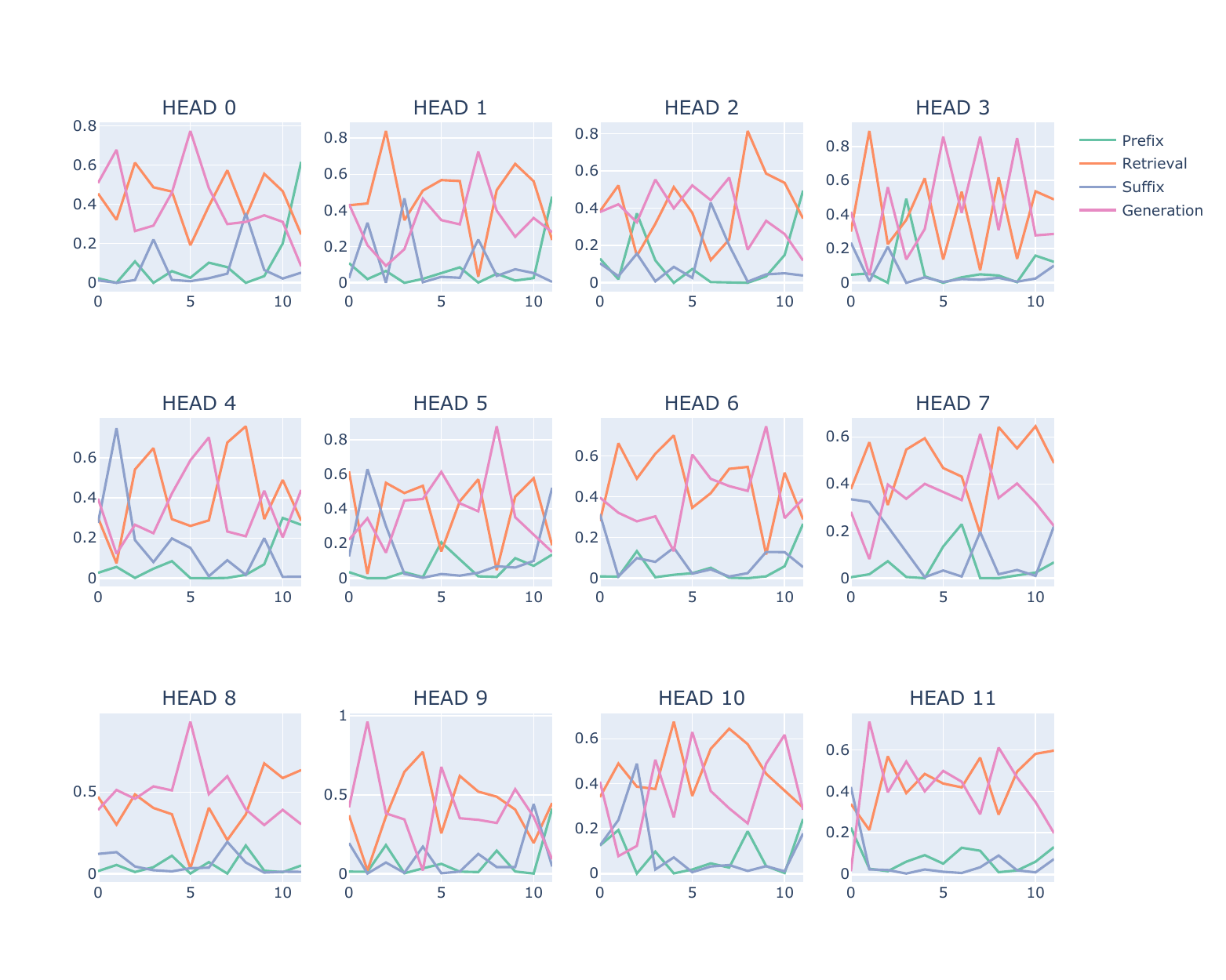}}\\
    \subfloat[\label{subfig:gpt2-cross-attention-cls}Cross attention distribution. Distribution of max attention scores of the interaction between two type of image patches (cls, others) and all text tokens. ]{\includegraphics[width=1\linewidth]{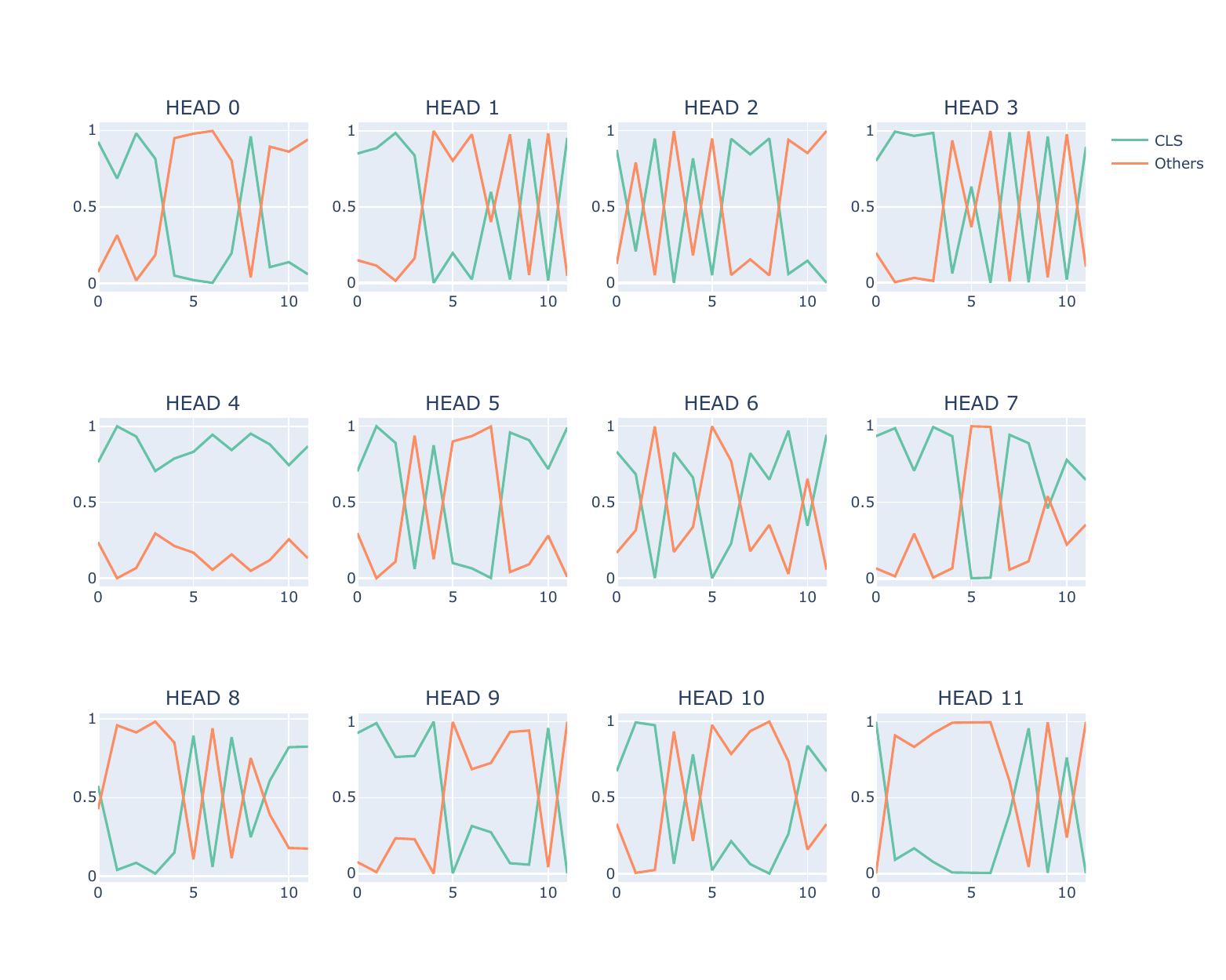}}
    \caption{Statistics of max attention scores in self and cross attentions from different different layers and heads with \textsc{SmallCap} (GPT2 variant). Compute the proportion of each attention scores from self and cross attention belongs to which parts.}
    \label{fig:gpt2-self-cross-attentions}
\end{figure}

\begin{figure}[ht]
    \centering
    \subfloat[\label{subfig:opt-self-attention}Self attention distribution.]{\includegraphics[width=1\linewidth]{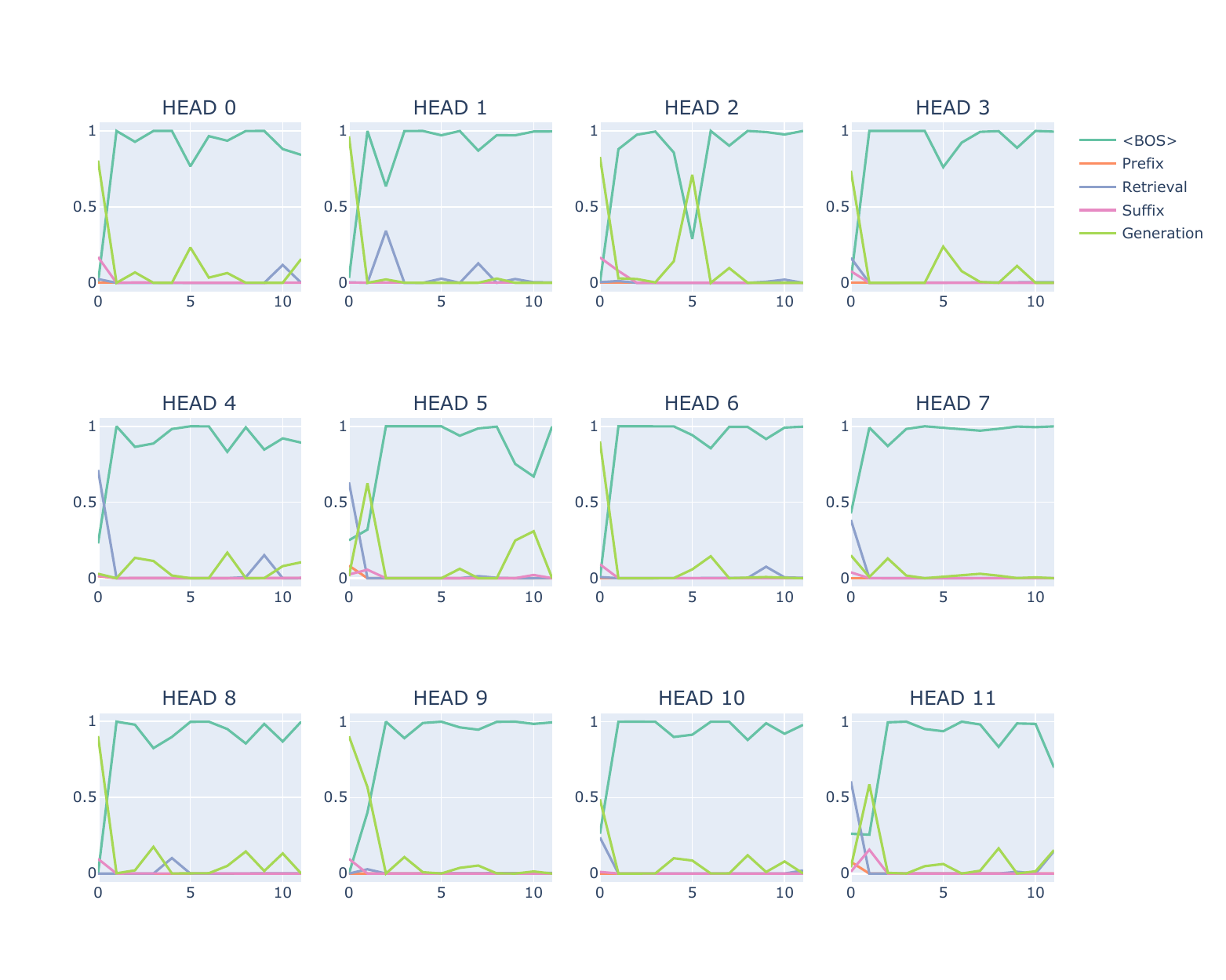}}\\
    \subfloat[\label{subfig:opt-cross-attention-text}Cross attention distribution. Distribution of max attention scores of the interaction between various part of text prompt and image patches.]{\includegraphics[width=1\linewidth]{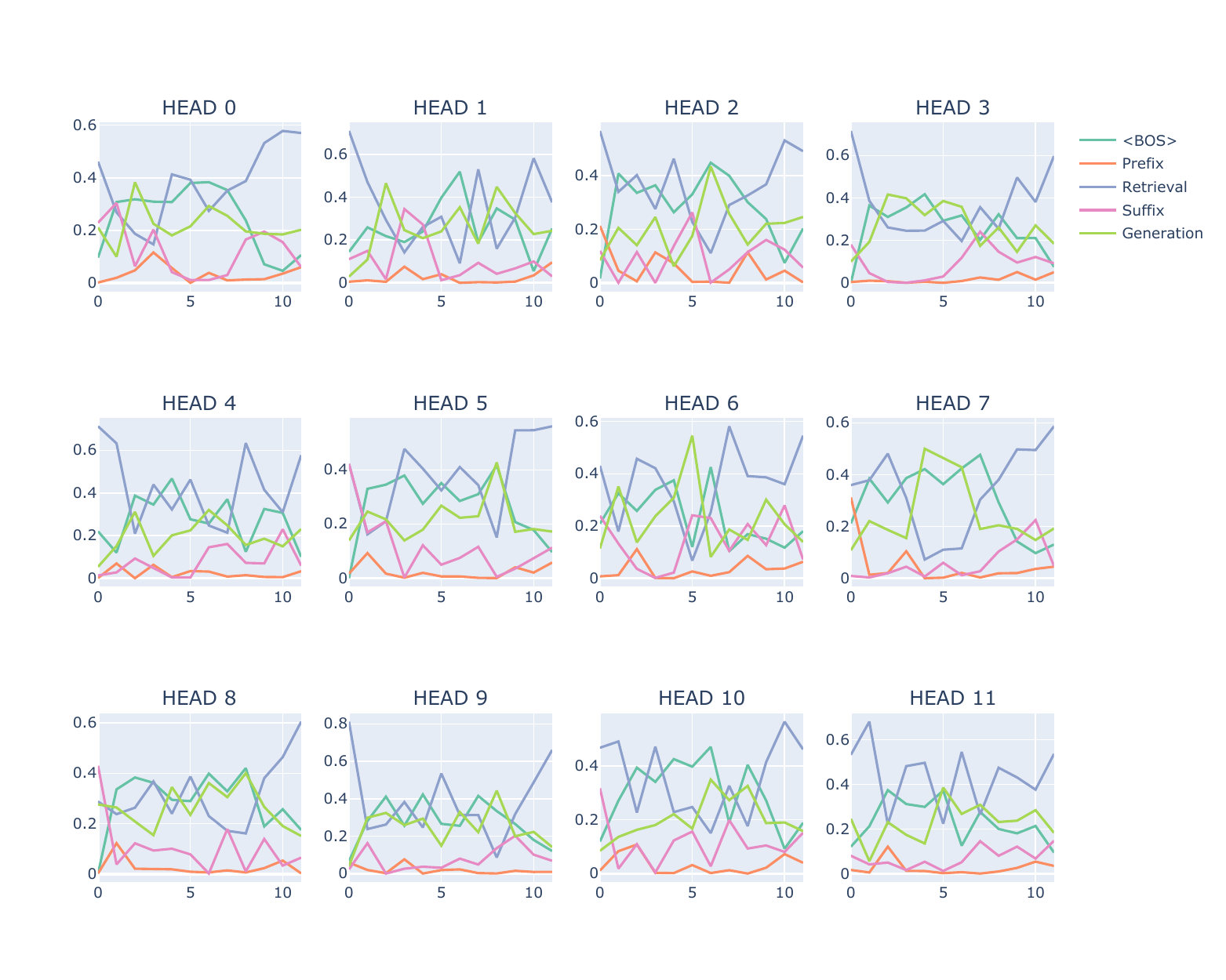}}\\
    \subfloat[\label{subfig:opt-cross-attention-cls}Cross attention distribution. Distribution of max attention scores of the interaction between two type of image patches (cls, others) and all text tokens. ]{\includegraphics[width=1\linewidth]{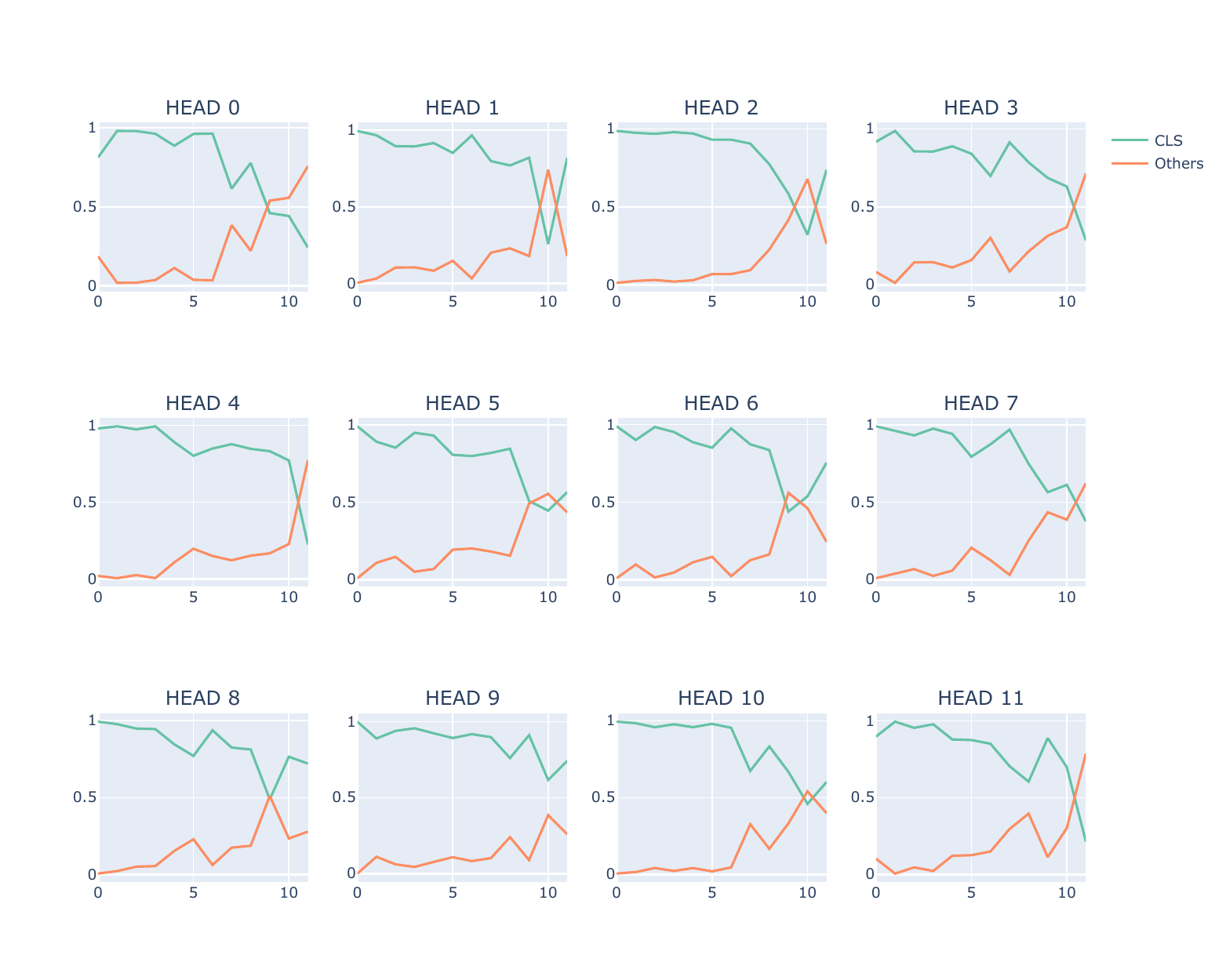}}
    \caption{Statistics of max attention scores in self and cross attentions from different different layers and heads with \textsc{SmallCap} (OPT-125M variant). Compute the proportion of each attention scores from self and cross attention belongs to which parts.}
    \label{fig:opt-self-cross-attentions_all}
\end{figure}

In Figure~\ref{fig:gpt2-self-cross-attentions} and Figure~\ref{fig:opt-self-cross-attentions_all}, we depict the distributions of both self-attention and cross-attention scores across various heads and layers for \textsc{SmallCap} with GPT-2 and OPT decoder variants.

\section{Qualitative examples}
\label{appendix:qua}
\begin{figure*}[ht]
    \centering
        \includegraphics[width=0.96\linewidth, height=2.8in]{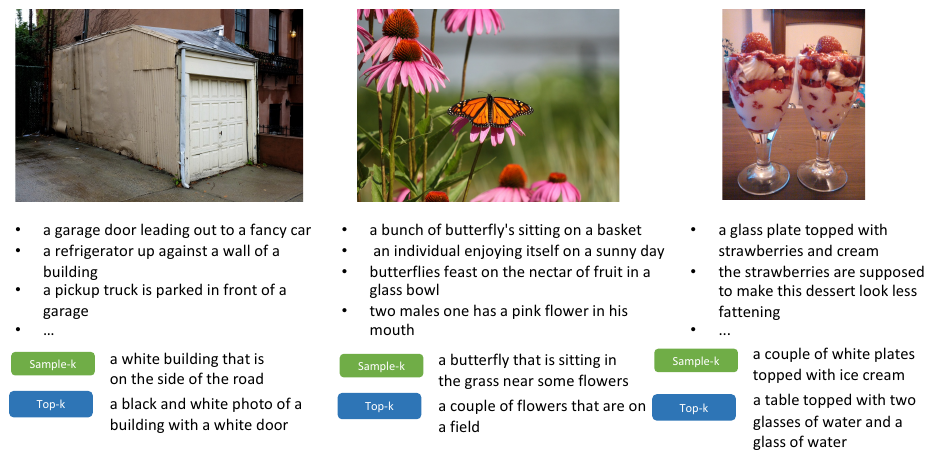}
        \caption{Qualitative examples of generated captions on NoCaps \textbf{out-domain} samples where the captions retrieved for the given image can be noisy and irrelevant. Here we retrieve four captions for each image. \label{fig:qua-nocaps}}
\end{figure*}

\begin{figure*}[ht]
    \centering
        \includegraphics[width=0.96\linewidth, height=2.8in]{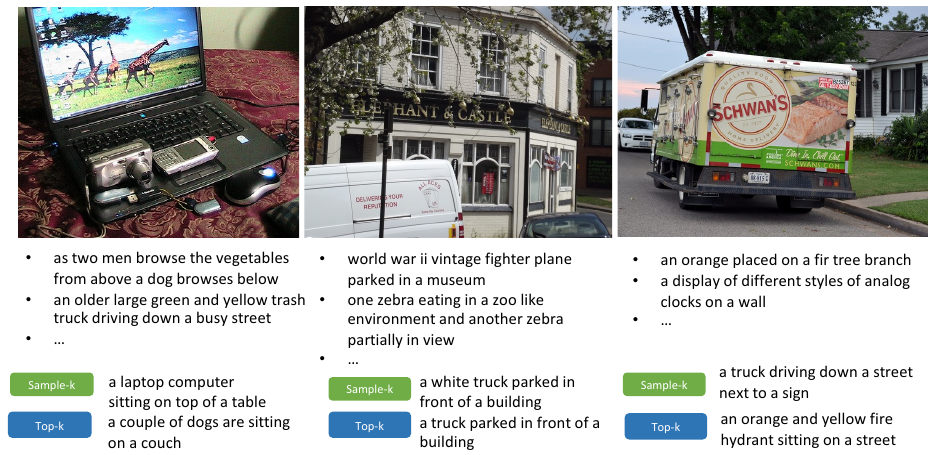}
        \includegraphics[width=0.96\linewidth, height=2.8in]{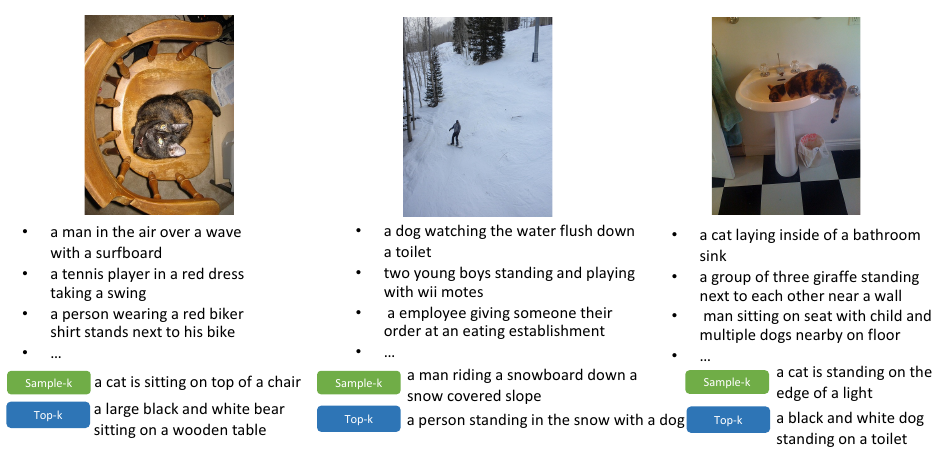}
        \caption{More qualitative examples of generated captions when \textbf{randomly} retrieving four captions for a given image. \label{fig:pred}}
\end{figure*}

We show more qualitative examples in Figure~\ref{fig:qua-nocaps} and Figure~\ref{fig:pred}.

\section{More results\label{appendix:more_results}}

\paragraph*{Order robustness evaluation}

In Table~\ref{tab:order-coco-ref} and Table~\ref{tab:order-nocaps-ref}, we provide both CIDEr and BLEU4 scores for order robustness evaluation (Section~\ref{sec:robust-eval}).
\begin{table}[t]
    \resizebox{\linewidth}{!}{
     \begin{tabular}{llrr}
        \toprule
        \multicolumn{2}{c}{Retrieval Order} & \multicolumn{2}{c}{SmallCap LM}\\
        Training & Evaluation & \multicolumn{1}{c}{GPT-2} & \multicolumn{1}{c}{OPT} \\
        \midrule 
        \multirow{3}{*}{default}
        & default & 116.4/36.1 &  120.3/37.1 \\
        & permute &  116.2/36.0 & 120.1/37.0 \\
        & reverse & 115.8/36.0	& 119.7/36.8 \\
        \midrule
        permute & permute &  \textbf{117.2/36.4} & \textbf{120.4/37.2} \\
        reverse & reverse & 116.4/36.1 & \textbf{120.7}/37.0	\\ 
        \bottomrule
        \end{tabular}
        }
    \caption{Results of manipulating the order of the top-k retrieved captions by either randomly permuting or reversing the list. We report CIDEr/BLEU4 scores on the COCO validation set using either a GPT-2 or OPT backbone in the SmallCap model. \label{tab:order-coco-ref}}

\end{table}

\begin{table}[t]
    \resizebox{\linewidth}{!}{
     \begin{tabular}{llrrr}
        \toprule   
        Model & Retrieval Order & In & Near & Out \\
        \midrule
         \multirow{2}{*}{GPT2} & default & 80.1/37.9 &79.4/35.9 &69.6/25.3 \\
         & permute &  \textbf{81.5/38.8} & 79.7/36.6 & \textbf{69.8/26.2} \\
         & reverse & 80.4/38.4	& \textbf{80.1/36.3}	& 68.4/25.1\\
         \midrule
        \multirow{2}{*}{OPT} & default & 91.0/27.1 & 84.4/23.8 &	76.3/15.0 \\
         & permute &  \textbf{94.2/28.6} & 84.0/25.0 & \textbf{79.4/15.8} \\
         & reverse & 92.5/28.4 & \textbf{85.6/25.3}	& 75.9/14.2\\ 
        \bottomrule
        \end{tabular}
        }
    \caption{Complete results with both CIDEr/BLEU4 on the NoCaps dataset when evaluated with different order of the top-four retrieved captions. The order applies to both train and evaluation stage.\label{tab:order-nocaps-ref}}
    \vspace{-12pt}
\end{table}

\paragraph*{Number of retrieved captions for sample-$k$ training}
We experiment with different size of the retrieval candidate list from which we randomly select captions for sample-$k$ training (Table~\ref{tab:largek}).

\begin{table}[ht]
\centering
    \resizebox{\linewidth}{!}{
     \begin{tabular}{llccccccc}
        \toprule   
          & & & \multicolumn{3}{c}{COCO} & 
          \\ \cmidrule(rl){4-6} 
        Size & $k$ & VizWiz & top-$k$ & last-$k$ & random\\
        \midrule
         7 & 4  
        & \text{36.0}
         & \text{119.2} & \text{117.1}	&\text{71.0}	
         \\
          10 & 4  
        & \text{36.0}
         & \textbf{119.3} & \textbf{118.3}	&\text{67.6}	
         \\
         50 & 4  
        & \text{33.9}
         & \text{118.1} & \text{117.7}	&\textbf{81.2}	
         \\
        \arrayrulecolor{black}\bottomrule 
        \end{tabular}
        }
    \caption{CIDEr score when sampling from different size of retrieval candidates. We see more improvements on random $k$ evaluation while almost keeping the same level of in-domain performance. With more noise involved during training, we would expect a longer training time would yield more robust performance.\label{tab:largek}}
\end{table}

\paragraph*{Percentage of tokens that are likely to be copied}
In Table~\ref{tab:mt-token-pct} we show the percentage of tokens that are likely to be copied from retrieved captions averaging through all samples in the validation set. Majority tokens takes more than half of the copied tokens.

\begin{table}[ht]
    \resizebox{\linewidth}{!}{
     \begin{tabular}{lrrrr} 
         & k=1 & 2 & 3 & 4 \\
        \toprule 
        \rowcolor[gray]{0.95}
        $T_{R}\in\text{Pred}$ & 49.1 & 63.3 & 69.8 & 75.7 \\
        $T_{R}\in\text{Pred (sample-k)}$  & 46.0 & 61.5 & 69.5 & 74.0 \\
        \rowcolor[gray]{0.95}
        $T_{M}\in\text{Pred}$  & - & 33.1 & 45.7 & 54.5 \\
        $T_{M}\in\text{Pred (sample-k)}$  & - & 32.5 & 45.3 & 53.3 \\
        \bottomrule
        \end{tabular}
        }
    \caption{Percentage of tokens in the predicted caption that are likely copied from majority tokens in retrieved captions in the COCO validation set. $T_{R}$ represent tokens in retrieved captions. $T_{M}$ represent the majority tokens in retrieved captions. \label{tab:mt-token-pct}}
\end{table}

\paragraph*{Comparison with other methods}
Inspired by~\citet{ret-robust23}, we have considered intentionally including less relevant captions by including one irrelevant caption, one low-ranked caption, and top-2 relevant captions instead of using top-4 retrieved captions. However, in our preliminary experiments, this strategy does not perform as well as the sampling approach, likely due to the high noise level it introduced.
\begin{table}[h]
    \centering
    \resizebox{\linewidth}{!}{
    \begin{tabular}{lcccccc}
        \toprule
        & \multicolumn{3}{c}{COCO Evaluation} & \multicolumn{3}{c}{NoCaps Evaluation} \\
        \cmidrule(lr){2-4} \cmidrule(lr){5-7}
        Method & top-k & last-k & random & In & Near & Out \\
        \midrule
        top-4    & 120.1 & 117.1 & 70.1 & 87.4 & 79.6 & 68.3 \\
        sample-4  & 119.2 & 118.6 & 73.1 & 89.7 & 80.9 & 71.1 \\
        mixed-4   & 119.2 & 118.1 & 66.7 & 59.9 & 57.9 & 39.4 \\
        \bottomrule
    \end{tabular}}
    \caption{CIDEr on COCO and NoCaps.\label{tab:other}}
\end{table}

\end{document}